# Non-planar 3D Printing of Double Shells

Ioanna Mitropoulou[1][0000-0001-7528-6216], Amir Vaxman[2][0000-0001-6998-6689], Olga Diamanti[3][0000-0003-2883-2194], Benjamin Dillenburger[1][0000-0002-5153-2985]

[1] Digital Building Technologies, ETH Zurich, Switzerland
[2] Institute of Perception, Action and Behaviour, The University of Edinburgh, UK
[3] Institute for Geometry of the Department of Mathematics, TU Graz, Austria.

**Abstract.** We present a method to fabricate double shell structures printed in transversal directions using multi-axis fused-deposition-modeling (FDM) robotic 3D printing. Shell structures, characterized by lightweight, thin walls, fast buildup, and minimal material usage, find diverse applications in prototyping and architecture for uses such as façade panels, molds for concrete casting, or full-scale pavilions. We leverage an underlying representation of transversal strip networks generated using existing methods and propose a methodology for converting them into printable partitions. Each partition is printed separately and assembled into a double-shell structure. We outline the specifications and workflow that make the printing of each piece and the subsequent assembly process feasible. The versatility and robustness of our method are demonstrated with both digital and fabricated results on surfaces of different scales and geometric complexity.

**Keywords:** Robotic 3D Printing, Non-planar, Strips, Digital Fabrication, Double Shell, Print Path Design, FDM, Digital Fabrication.

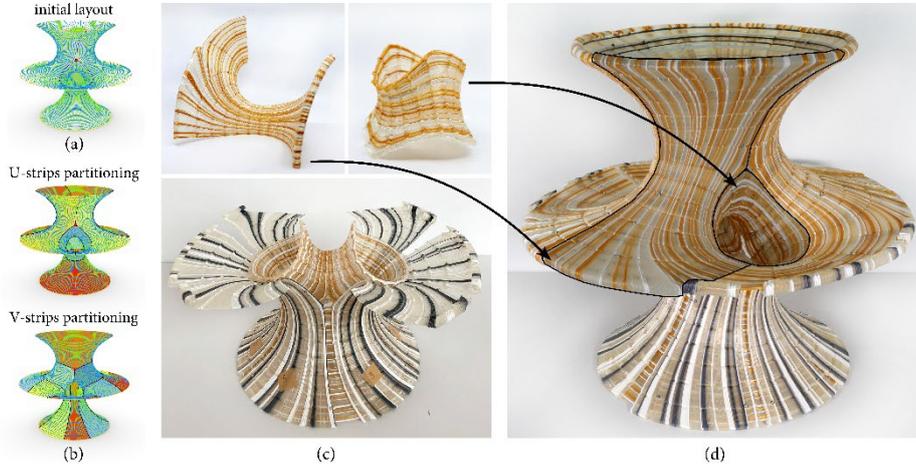

**Fig. 1.** FDM robotic non-planar 3D printing of double shells along transversal directions. Fabrication of the Costa minimal surface using paths aligned with principal curvature directions. (a) Input Strip-Decomposable Quad (SDQ) mesh. (b) Fabrication-aware partitioning for the $U$ (top) and the $V$ (bottom) strip networks, each partition is shown with a different color. (c) Printing of individual pieces and assembly. (d) Final prototype.

## 1   Introduction

Latest developments in Additive Manufacturing (AM) have opened new possibilities for 3D printing objects with unprecedented geometric complexity. We focus on fused deposition modeling (FDM), whereby a linear molten thermoplastic material is extruded from a hot nozzle into layers that accumulate to produce an object. FDM is relatively inexpensive and flexible (Gao et al. 2015) and has gained much popularity among industrial and hobbyist practitioners.

The introduction of robotic arms in FDM printing enables the scaling up of the printing process. In larger-scale printing, material deposition often takes the form of a hollow shell, where each layer is realized by a single path without multiple offsets or infill. Shells have the advantage of less material usage and faster buildup, and thus are used for various architectural applications such as façade panels (Sarakinioti et al. 2018; Taseva et al. 2020), bespoke floors (Aectual 2019), and molds for concrete casting (Jipa et al. 2017; Leschok and Dillenburger 2019; Burger et al. 2020).

However, the realization of single-layer FDM shells has some inherent disadvantages. First, the thickness of the shell is constrained by the size of the extrusion nozzle. For FDM printing using filament extrusion, this is typically between 0.5 and 2.5 mm, which can cause instabilities when scaling up. Second, due to the one-dimensional nature of the paths, there is an asymmetry in the quality of the approximation of the surface: the direction parallel to the paths is smoothly approximated, while the direction orthogonal to the paths is only approximated with a staircase of subsequent sections (**Fig. 2**a). This adversely affects both aesthetics, with a staircase effect in the direction orthogonal to the paths, and structural strength, as the resulting object is anisotropic, with considerably higher strength along the print direction (Ulu et al. 2015). While both the staircase effect and anisotropy are hard to avoid since they are inherent to the printing process, they are properties that can be controlled to improve the strength and quality of the printed object.

To that end, we propose the design of double (**Fig. 1**, **Fig. 2**c) instead of single (**Fig. 2**b) shells, with two outer walls printed using non-planar paths in transversal directions and connected with sparse links (ribs) along a hollow interior. This adds structural depth to the resulting object, thus improving stability. It also allows the distribution of imperfections more evenly along different directions on each side of the shell.

## 2 Related work

### 2.1 Non-planar slicing and multi-axis printing

Traditional planar 2.5D printing (Gibson, Rosen, and Stucker 2009) is based on the accumulation of planar layers along a single build direction. Multiple approaches for non-planar printing still follow the layered approach but allowing for curved layers. In perhaps the first work in that direction (Chakraborty, Aneesh Reddy, and Roy Choudhury 2008), the printer can dynamically change $z$-values within individual layers. Since then, multiple works have pursued the direction of decomposing a shell or volume into curved layers (Ahlers et al. 2019; Allen and Trask 2015; Chen et al. 2019; Etienne et al. 2019; B. Huang and Singamneni 2015; Lim et al. 2016; Llewellyn-Jones, Allen, and Trask 2016; Pelzer and Hopmann 2021). Curved slicing produces non-planar print paths that can greatly reduce support, improve surface quality (in particular, reduce staircase artifacts), and improve mechanical properties.

To facilitate the fabrication of non-planar paths, one can use a robotic arm with more degrees of freedom (DOF) compared to the standard triaxial system of traditional desktop printers (Keating and Oxman 2013; Pan et al. 2014; C. Wu et al. 2017). Multi-DOF systems have been used to print wireframe models (Y. Huang et al. 2016; R. Wu et al. 2016). They have also been used for layered printing; for example, Dai et al. (Dai et al. 2018) use a multi-DOF system that keeps the printer head fixed and freely moves the object during printing to print curved layers that eliminate sacrificial support. However, moving the object during the fabrication process can be challenging when printing larger objects. Moving the print tool instead of the object is an alternative used in various research projects (Mitropoulou, Bernhard, and Dillenburger 2020; Fang et al. 2020a; Zhang et al. 2022).

### 2.2 Fabrication in transversal directions

An important challenge of FDM 3D printing is the anisotropic behavior of printed objects due to the weak adhesion between layers along the vertical axis (Ulu et al. 2015; Fang et al. 2020b). In response to this challenge, researchers have explored alternative printing paths using transversal directions, such as aligning print directions with stress lines (Tam and Mueller 2017), also integrating carbon fiber (Kwon et al. 2019). Utilizing transversal directions to enhance fabrication isn't confined to just 3D printing. This approach is prevalent in fabrication processes that involve linear elongated segments, where the discontinuous direction has weaker bonding than the continuous one. Examples include crafting surfaces with paper strips (Takezawa et al. 2016) or metal rods (Ma et al. 2020). The key advantage is that when directions intersect, the strengths in one direction compensate for the vulnerabilities in the other, resulting in a more robust final product.

### 2.3 Strips

Utilizing strips to represent the underlying geometry is common in such fabrication processes because the geometric characteristics of strips match the elongated fabrication units. Strips have been used to design single-curved façade panels (Pottmann et al. 2008), textile ribbons (Schüller, Poranne, and Sorkine-Hornung 2018) and developable surfaces (Verhoeven et al. 2022).

In the same spirit, (Mitropoulou et al. 2024) design transversal strip layouts overlayed into a Strip-Decomposable Quad (SDQ) mesh (**Fig. 2**d), considering fabrication-related properties and user-controlled alignment. They propose using SDQ meshes as an intermediate representation to describe non-planar 3D printing paths. Our work follows that paradigm; however, we take advantage of both transversal strip networks to fabricate double instead of single shells printed along transversal directions. Further, we extend this scheme with a partitioning method that guarantees both topological and geometric feasibility for printing.

## 3 Overview

We begin with a set of two transversal strip networks, $U$ (depicted in blue) and $V$ (depicted in green), combined in an SDQ mesh (**Fig. 2**d). The novelty of this work consists of considering each strip network to represent the non-planar print paths of each side of a double shell, printed in transversal directions. The challenges we tackle to enable this fabrication mode are the following.

— Partitioning strip networks into printable patches that can be fabricated with our print setup (Sections 4.1, 4.2).
— Addition of fabrication-enabling details and assembly into one connected double shell object (Section 4.3).

### 3.1 Paths representation

Paths are generated by subdividing $N$ times each strip $S_i$. Unlike planar printing, here, the print direction $T$ (i.e., the orientation of the robot's print head while printing) and the layer height $h$ are variable. **Fig. 2**e shows how paths are generated on a strip $S_i$ and the resulting $T$ and $h$ for two points $a$ and $b$. $T$ is given by the direction of the transversal (green) edge, and $h$ is given by the distance from the neighboring subdivision.

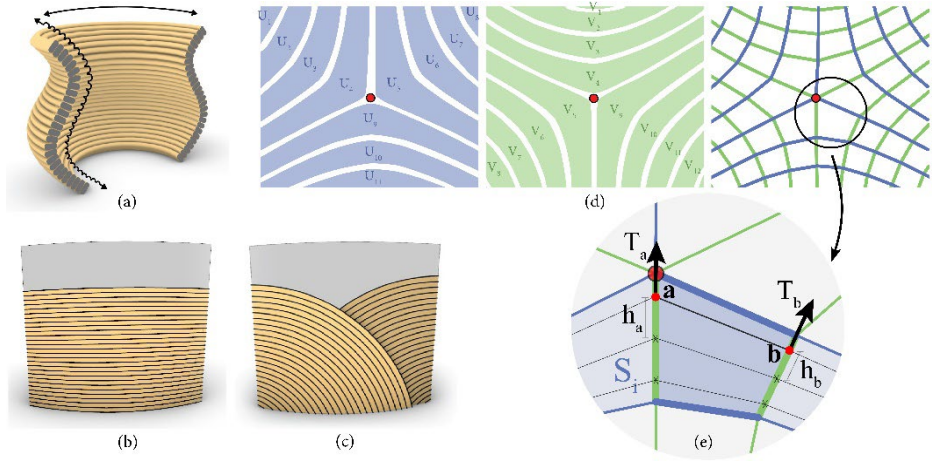

**Fig. 2.** (a) In 3D printed surfaces, one direction is smoothly approximated, while the other is approximated with a staircase of subsequent sections. (b, c) Diagram of a single and a double shell with each side printed in transversal directions. (d) Two transversal strip networks, $U$ (left) and $V$ (middle), overlayed into one SDQ mesh (right), illustraton from (Mitropoulou et al. 2024). (e) Paths generation by subdividing a strip $S_i$ showing the layer height $h$ and print direction $T$ for two points $a, b$.

### 3.2  Printing standing shells

Next, we describe our non-planar 3D printing setup of standing shells (**Fig. 3**, top), i.e., shells where only the first path lies on sacrificial support, and all subsequent paths lie on previously deposited paths. The robotic arm orients the extruder (i.e., the print tool) towards the print direction $T$ at each position and prints sequentially on a sacrificial base, from first to last path, in an uninterrupted motion. This sequential logic requires a print sequence, where each path being printed fully lies on previously deposited paths. Since the layer height $h$, i.e., the distance between consecutive layers, varies, the extruder updates the material flow rate at each print position to match the quantity of material necessary for filling the current layer height. On the other hand, the layer width $w$ is considered constant throughout the process.

The robot's degrees of freedom provide considerable flexibility for reorienting the tool around the object and depositing material in various orientations. However, the range of possible orientations for material deposition isn't unlimited. During the buildup of paths, print angles that are unfavorable for material deposition can appear (**Fig. 3**, bottom right). For that, we set a maximum print angle variation $\gamma$ that should not be exceeded (for our setup $\gamma = \pi/2$). In addition, the robot cannot print outside of its reachable space; we define a bounding box of reachable space (for our setup, it has dimensions 50 x 50 x 50 cm), and all print geometry must lie within it.

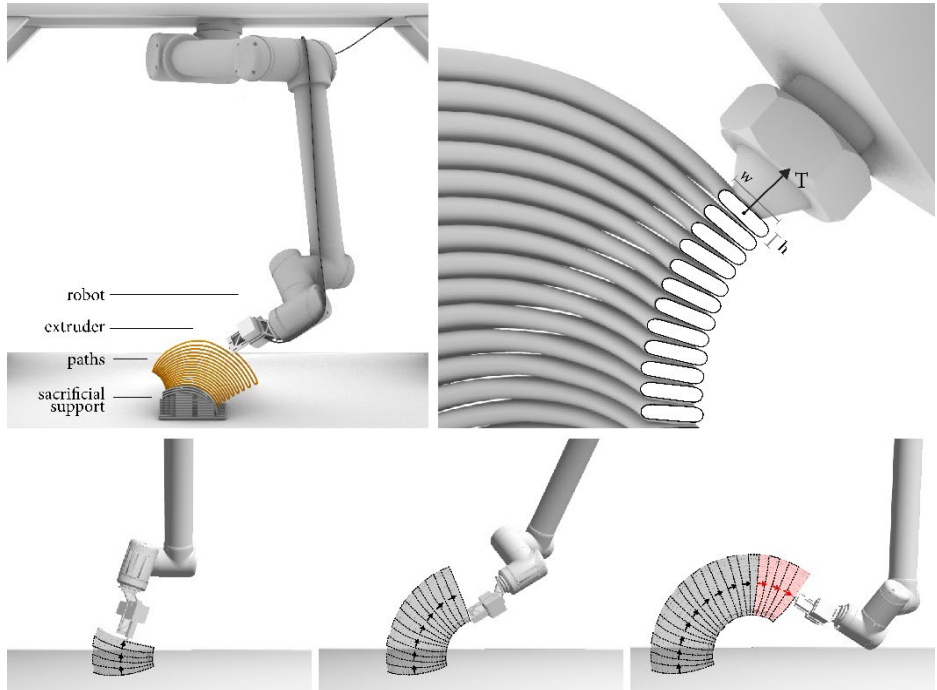

**Fig. 3.** Top: Overview of a standing shell's robotic 3D printing process. Bottom: Print angles during the buildup of paths can become unfeasible (right, in red) if the print direction varies a lot.

**Partitioning constraints**

Considering the characteristics of the process outlined above, we propose a partitioning scheme of strips that considers the following constraints.

1. *Topologically feasible*. All patches must have a topologically simple sequence of paths so that there is an unambiguous print sequence. This should consist of a single non-branching sequence of curves that layer upon each other so that each path being printed lies on a previously deposited path.
2. *Geometrically feasible*. For each patch, the dimensions must fit within the bounding box, and the angle variation should not exceed the maximum print angle variation.
3. *Non-overlapping seams*. The partitioning cuts on the two strip networks $U$ and $V$ must not overlap so that upon assembly, the two shells form one connected object.

### 3.3  Workflow

Our workflow proceeds as follows.

– Begin with an SDQ mesh.

- Carry out **topological partitioning** that separates the strip networks into simply-connected patches (Section 4.1).
- Carry out **geometric partitioning** into patches that respect the angle and size bounds (Section 4.2).
- Incorporate **assembly-related details** and compute fabrication data (Section 4.3).
- Print and assemble the individual pieces to obtain the final output (Section 5).

## 4 Method

### 4.1 Topological partitioning

Topological partitioning (**Fig. 4**) aims to separate the strip networks $U$ and $V$ into simply-connected patches so that each patch has an unambiguous ordering (up to orientation) and can be printed as a standing shell.

In (Mitropoulou et al. 2024), a partitioning scheme is detailed that cuts the strip network along the separatrices of the singularities and separates handles, thus splitting a strip network into simply-connected parts, as shown in **Fig. 4**a and **Fig. 4**d. These cuts continue until they form closed loops or terminate on open boundaries. The difference here is that we aim to partition both $U$ and $V$ (i.e., both **Fig. 4**a and **Fig. 4**b, similarly both **Fig. 4**d and **Fig. 4**e), and we have the additional constraint that the cuts of the two networks should not overlap. As long as the partitioning is parallel to the direction of strips (i.e., no strip is cut perpendicularly), then the cuts of $U$ and $V$ are non-overlapping, as they are applied in transversal directions, for example, see the partitioning of the D6 singularity (**Fig. 4** a, b, c).

However, around the D2 singularity (**Fig. 4** d, e, f), cutting along the separatrix parallel to the strips alone would not separate the self-incident strip at the cut, which creates paths unfeasible for printing due to their self-obstructing geometry. As a result, around the D2 singularity, topological cuts are added that are both parallel to the strips (**Fig. 4** d, e, in black) and intersect them transversally (**Fig. 4** d, e, in red). This causes an overlap of cuts as, essentially, the cuts of the $U$ and the $V$ network are identical. If those cuts were allowed to terminate naturally, they would create an extensive sequence of edges where both sides share a cut. Then, upon assembly, the final object would be disconnected there. To avoid this, we introduce two additional transversal cuts $C_U$ and $C_V$ in distance $d_q$ edges from the singularity, that interrupt the overlapping cut (the resulting reduced overlap is marked in cyan). As a result, the pieces are still connected upon assembly.

### 4.2 Geometric partitioning

We next partition the patches further to adhere to the sizing and angle constraints (**Fig. 5**). To avoid overlapping cuts in this step, we keep track of the edges that participate in cuts on each network and disallow adding overlapping cuts. Since every edge sequence of the mesh can potentially become a cut, there are always more than enough edges to use as cuts despite this restriction. The geometric partitioning cuts are interrupted at the boundary of each patch; thus, patches can be partitioned independently.

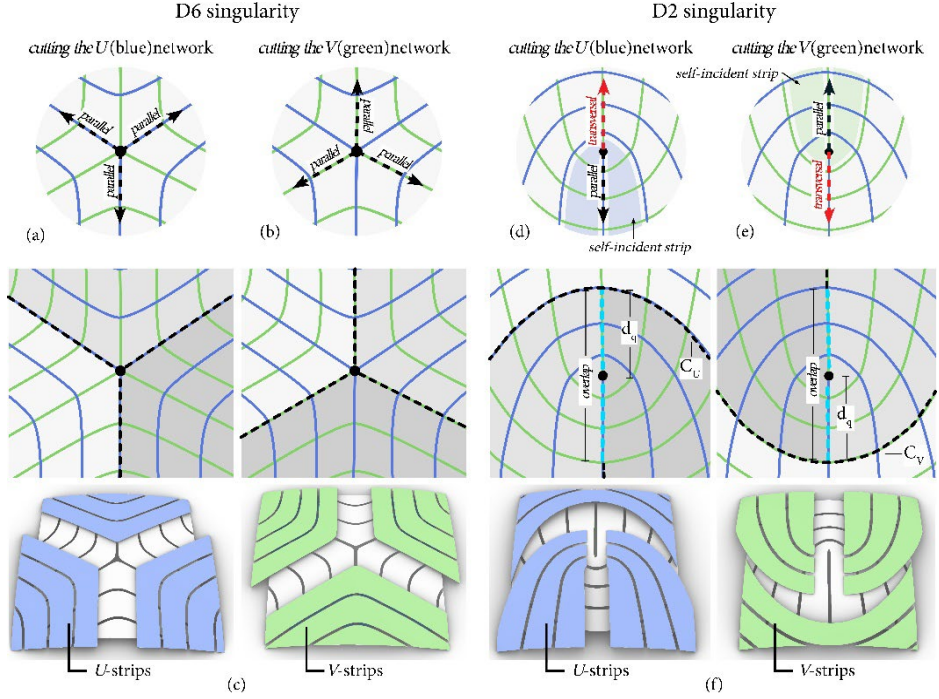

**Fig. 4.** Topological partitioning of the two singularities most commonly found in SDQ meshes for both $U$ and $V$ strips. The strips' outline is displayed as smooth for easier comprehension. (a, b, c) D6 singularity partitioning along separatrices. (d, e, f) D2 singularity partitioning along separatrices. The cyan line illustrates the overlapping cut on the two sides, bounded by the transversal cuts $C_U$, $C_V$ in distance $d_q$ edges from the singularity.

**Partitioning to fit size bounds.**
In each patch, we find the two transversal strips $S_U$ and $S_V$ with the largest area. In addition, we compute the maximum extent of each patch using Principal Component Analysis (PCA). Each resulting principal direction is best aligned with either $S_U$ or $S_V$. If a direction exceeds the maximum size, we partition the according strip by adding the appropriate number of equidistant cuts (in number of quads), ignoring edges that would cause overlap. The resulting partitions are checked again, and the process is repeated until no piece exceeds the set bounds. **Fig. 5** illustrates the partitioning of the $U$ and $V$ networks of a doubly-curved surface that is, by design, larger than the bounding box.

**Partitioning to fit angle bounds.**
While the robot is printing with the extruder oriented in the print direction $T$, its orientation change should not exceed a certain threshold $\gamma$. Note that the angle with which the robot is printing depends on the piece's orientation on the build platform. This is why we are not examining angles with the gravity direction (these are defined later when the piece's print orientation has been decided) but rather angle variations, which are independent of the piece's orientation on the print platform.

Assume we are working on a patch of $U$ strips; its $U$-strips (blue) lie along the paths' direction, and its $V$-strips (green) are transversal to it. As a result, the total angle variation $A = \sum |\alpha_{V_i}|$ (the sum of absolute angle differences of neighboring $T$) within $V$-strips measures the variation during the buildup of the piece. We consider the V-strip with the greatest angle variation $A_{max}$. If $A_{max} > \gamma$, then we cut the strip with $k = ceil(A_{max}/\gamma)$ equidistant (in number of quads) cuts, again ignoring edges that would cause overlap. All resulting patches are rechecked, and the process is repeated until no patches have $V$ strips that exceed the $A_{max}$. We show the process in **Fig. 5**, using different angle thresholds $\gamma$.

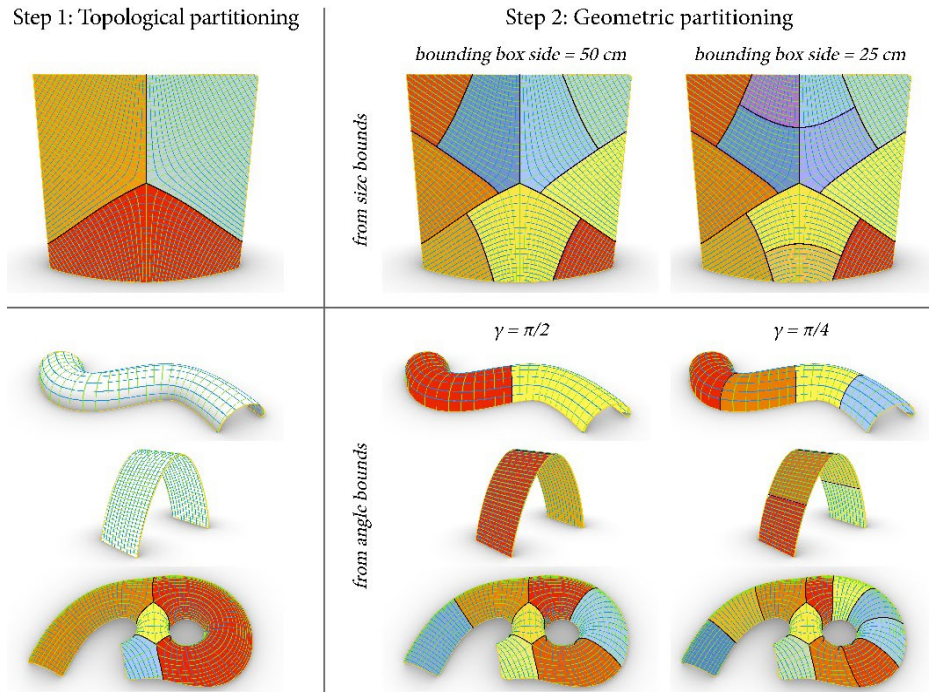

**Fig. 5.** Partitioning steps: first, topological partitioning (left), and second, geometric partitioning (right). Top: Partitioning a doubly curved surface from size bounds using two different bounding box sizes. Bottom: Partitioning various surfaces from angle bounds using two different γ thresholds. In both cases, only the $U$-strips partitioning is displayed.

### 4.3   From digital to physical

In the following, we provide the steps for preparing the digital double shell model to be ready for printing.

**Surface thickening – offsetting**

The double-shell surface consists of two shells printed along transversal directions and a hollow interior with sparse ribs connecting the shells. The total thickness $t$ of the shell can be determined by the user, where in most prototypes, we use the default value $t = 4n$ (**Fig. 6**a), where n is the nozzle diameter (2.5 mm in our setup). To account for the thickness of the shells, we offset the original surface on each side with default distance $t/2 - 0.5n = 1.5n$ in the direction of the normals of the mesh, which we compute for each quad by triangulating it and finding the average normal of its triangles. This offsetting scheme is adequate for our application since our shapes have low surface detail, and the quad mesh has sufficiently high resolution.

**Managing inaccuracies during assembly**

The assembly of the printed pieces might be hampered by small inaccuracies present in the prints. They originate either from inevitable process imperfections or minor deformations of the pieces during and after the print due to uneven shrinkage while cooling. To account for them, we strategically introduce tolerances by removing a single strip on each boundary between patches. We do this only on one side (by default, the $V$ side, **Fig. 6**b) to avoid creating holes that would otherwise appear at the intersection of gaps. For small inaccuracies, this strategy is sufficient to compensate for errors on both sides of the shell.

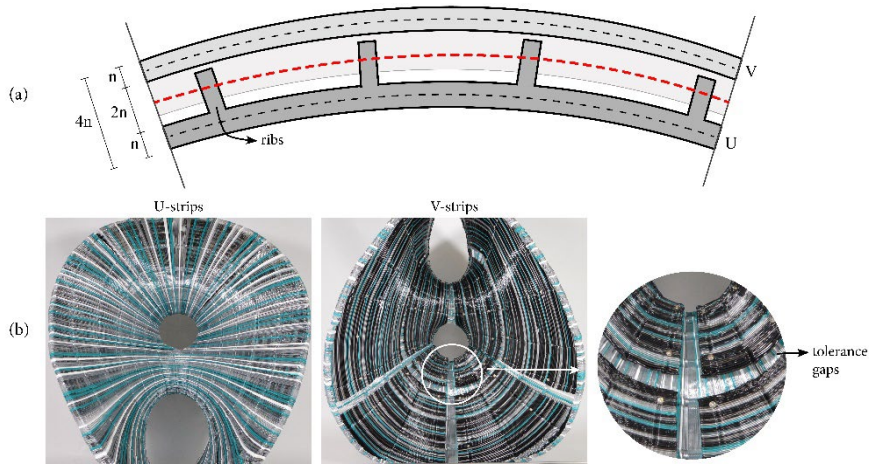

**Fig. 6.** (a) Offsetting the original surface (in red dotted line) on both sides to create a double shell of thickness $4n$. (b) Accounting for tolerances by creating one side with gaps (right, $V$-strips).

**Orienting for fabrication**

To decide on the global print orientation of a piece, we calculate the average $\vec{h}$ of all directions $T$ (**Fig. 7**a, left, in black). We also calculate the average $\vec{m}$ of the first strip's $T$ directions (**Fig. 7**a, left, in red). Then, the piece's global orientation is calculated so that the vector $0.5(\vec{h} + \vec{m})$ is aligned with the vertical $\vec{z}$ direction (**Fig. 7**a, right).

**Sacrificial support**

Once the pieces have been oriented for fabrication, a supportive base is created upon which the piece can be printed. It consists of a scaffolding, a lighter lower part produced by a liner hatch to save time and material, and a platform, an upper part, which connects to the printed object and consists of layered offsets of the first path of the patch (**Fig. 7**b).

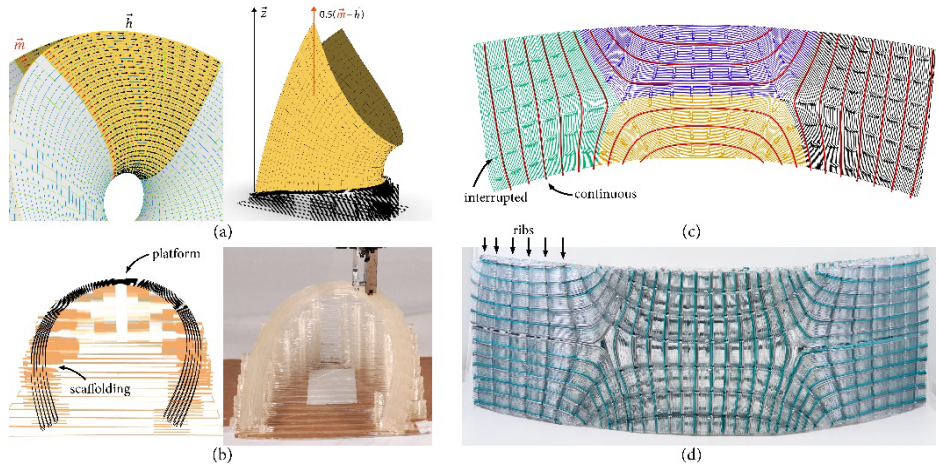

**Fig. 7.** (a) Calculation of orientation for fabrication so that $0.5(\vec{m} + \vec{h})$ is aligned with the vertical direction $\vec{z}$, where $\vec{h}$ is the average of all directions $T$ (vectors in black), and $\vec{m}$ is the average of the first strip's $T$ directions (vectors in red). (b) Sacrificial base diagram (left) and printing (right). (c) Rib interruptions for interlocking the two sides of the shell. The continuous ribs of one wall are denoted with red lines. The ribs on the other side (colored per patch) are interrupted to create an interlocking. (d) Printed prototype.

**Ribs and connections**

Along each wall, we print rigidifying ribs, which help with the overall stability and link the two sides of the shell upon assembly. The ribs follow the direction transversal to the one we are printing and have a default distance of 4 strips (chosen for aesthetics). Similarly to the print paths, the ribs intersect transversely on the two sides of the shell. To create an interlocking connection between the two sides, we interrupt the ribs of one side on the intersections with the ribs from the other side (**Fig. 7** c, d). This provides a guide for positioning the pieces during assembly while allowing minor adjustments by sliding, as we leave an additional tolerance gap between the ribs at their intersections. The pieces are kept in place using screws on rib intersections (visible in **Fig. 6**b right), which are drilled manually. The assembly process (displayed in **Fig. 9** for prototypes d and e) consists of assembling one piece from each side in an alternating pattern at every step so that each new piece is half attached to the existing structure.

**Printing process**

Our printing setup consists of a universal robot UR10 mounted overhead to gain better access to all sides of the printed object (**Fig. 8**). The plastic extruder tool has a nozzle with a diameter of 2.5mm and is controlled by an Arduino Mega board. The printing process is carried out at a constant linear robot speed at the extrusion point (15 mm/sec while printing the object and 23 mm/sec while printing support). We vary the extrusion rate to match the volume of material required to achieve the desired path thickness at every printing position. The change of filament color is carried out manually without stopping the printing process, and its purpose is solely aesthetic, to emphasize the path orientations and differentiate the two shells visually.

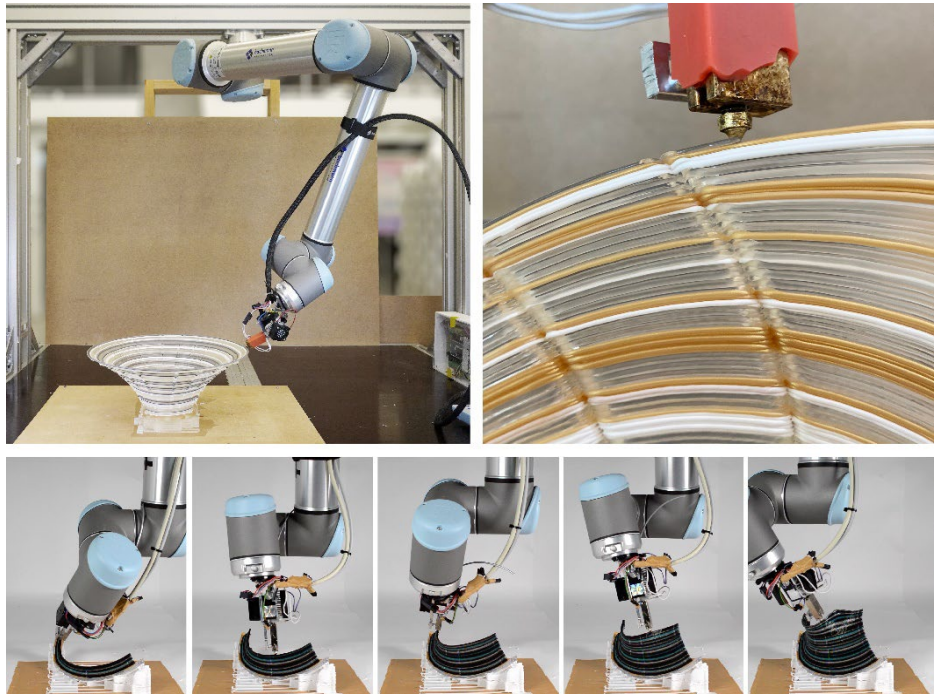

**Fig. 8.** Top: Robotic setup for fabrication. Bottom: timelapse of printing, total time 35 min.

## 5    Results

We present a variety of printed prototypes created with the described methods. **Fig. 1** presents the Costa minimal surface with the full pipeline from the initial input SDQ mesh to the final fabricated result. The shape is partitioned into 7 pieces on one side and 11 on the other. The prototypes shown in **Fig. 9** a, b, c are created with an SDQ mesh aligned to user-drawn directional constraints. The curved surfaces (**Fig. 9** a, b) are fabricated with 3 pieces on each side, while the elongated curved surface (**Fig. 9**c) is partitioned into 5 and 6 pieces. The Chen Gackstatter minimal surface (**Fig. 9**d) is generated using boundary-aligned constraints; the paths of one side are constrained to be parallel, and the paths of the other side to be orthogonal to the boundary. It is segmented into 5 and 16 pieces. Finally, the Batwing minimal surface (**Fig. 9**e) is generated with curvature-aligned directional constraints and is segmented into 9 and 12 pieces. **Table 1** presents fabrication details for all the prototypes.

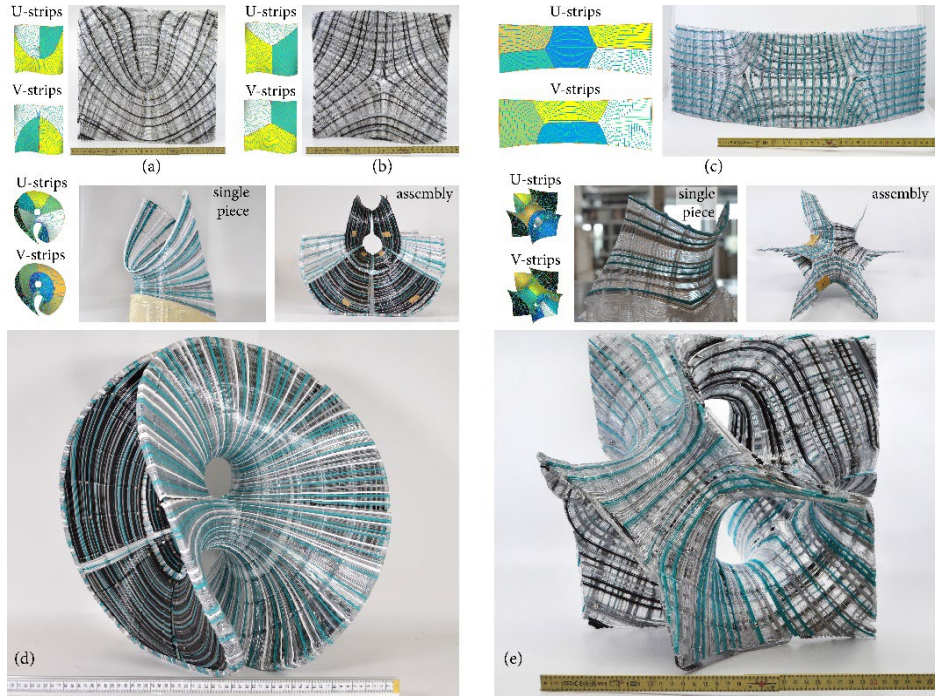

**Fig. 9.** (a, b) Curved surfaces with paths aligned to two different user-drawn directions. (c) Elongated curved surface with paths aligned to user-drawn directions. (d) Chen Gackstatter minimal surface with paths aligned to boundary directions. (e) Batwing minimal surface with paths aligned to curvature directions. More details on the prototypes are displayed in **Table 1**.

| model | Dimensions (cm) | #sings | #geometric cuts | # pieces $U$ | # pieces $V$ | Total print time (hrs) | % sacrificial support |
|---|---|---|---|---|---|---|---|
| **Fig. 1** | 60 x 60 x 51 | 4 | 12 | 7 | 11 | 51.5 | 49% |
| **Fig. 9a** | 30 x 6 x 30 | 1 | 0 | 3 | 3 | 9.6 | 52% |
| **Fig. 9b** | 30 x 6 x 30 | 1 | 0 | 3 | 3 | 8.9 | 48% |
| **Fig. 9c** | 55 x 17 x 10 | 1 | 0 | 5 | 6 | 8.1 | 54% |
| **Fig. 9d** | 50 x 50 x 60 | 4 | 8 | 5 | 16 | 46 | 42% |
| **Fig. 9e** | 30 x 30 x 30 | 7 | 4 | 9 | 12 | 16.8 | 28% |

**Table 1.** Details of prototypes presented in **Fig. 1** and **Fig. 9**.

# 6  Conclusion

Our pipeline successfully produces quality print paths for various shapes. We note some limitations and challenges.

*Challenges of double shells.* Despite the merits of printing double shells in transversal directions, this fabrication mode also has notable challenges. Printing a double shell implies that the surface is essentially printed twice, which leads to more pieces, more sacrificial support for their base, and thus longer fabrication times. Further advancing the partitioning method to consider the material efficiency of support structures can help reduce the waste material. Also, the printing in transversal directions takes place in separate pieces that are held together with sparse point links of small surface area (screws). As a result, the connection of transversal directions is significantly weaker compared to other approaches that print both directions directly on each other (Tam and Mueller 2017; Kwon et al. 2019).

*Assembly.* A well-known challenge in FDM, which our work is also susceptible to, occurs in the assembly of the printed paths. Minor inaccuracies caused by imperfections of printing, such as over-extruding or deformations of the pieces during cool-down, may cause them not to fit well together. When printing double shells, these can be especially problematic as the interfaces between pieces have a very large area; essentially, the entire pieces are interfaces, meaning that any imperfection can impair the assembly process. Future work into designing smarter part connections, such as puzzle joints or joints that take advantage of the double-shell thickness of the components, could prove very beneficial to the final quality of the assembled 3D-printed object.

*Offsetting.* A topic worth further study is how to offset shells optimally; our normal-based offsetting only offers an approximate solution and may need to be adapted for more intricate shapes.

## 6.1  Outlook

This work opens the door to a variety of promising future applications.

*From multi-color to multi-material printing.* We use colors to differentiate the curved paths visually. Such changes in the extrusion's content can have significant future applications. Printing different materials, instead of just different colors, can create functional material gradients that serve specific performance goals. For instance, fiber-reinforced filament can be used strategically to improve the resulting piece's structural properties. The development of specialized hardware that feeds the correct material at the correct times during the print would contribute to such controlled functional material changes.

*Applications in architecture.* 3D-printed shells in architecture find various applications but are often limited by their low thickness and anisotropy, which can be effectively mitigated through the use of double shells printed using non-planar layers. Printing each shell side separately and assembling them afterward is beneficial for embedding functional equipment like tubes or wires and easily accessing them for repairs. Also, it facilitates the incorporation of reinforcement between the two shells, enhancing the overall functionality and durability of the shell structure. Further, add-on printing is a key application of non-planar shell printing, enabling direct material deposition onto existing elements to improve efficiency, promote reuse, and facilitate repair or reinforcement. Structural optimization is another important application. By exploiting the anisotropic properties of 3D printed materials, objects can be optimized for specific load cases. Printing each side of the shell with a different orientation allows further control over the anisotropic properties, leading to additional mechanical improvements.

### 6.2  Reflection

Our research on the non-planar 3D printing of double shells using multi-axis FDM resonates with the conference's theme, "Beyond Optimization," challenging traditional approaches in robotic fabrication. We address this theme through two key aspects:

*Embracing real-world complexity in assembly tasks.* An important part of our assembly method is the acknowledgment and proactive management of inaccuracies. Unlike traditional approaches that strive for hyper-accurate results – often a challenging and resource-intensive endeavor – our method anticipates and accommodates inaccuracies by incorporating gaps in the seams for handling tolerances. In the real-world conditions of fabrication, absolute precision is often unattainable. This acknowledgment of fabrication imperfections demonstrates a shift from pursuing unfeasible precision to developing robust and adaptable systems that do not fail when unavoidable imperfections occur.

*Broadening the scope beyond the technical.* Our research presents a novel way for printing double shells using non-planar paths. Beyond functional efficiency, this control opens avenues for artistic expression and aesthetic enhancement. By manipulating print paths, we unlock new design possibilities, creating textures and forms that enhance both the visual and tactile qualities of the final product, which can be particularly valuable in artistic applications.

## 7  References

Aectual. 2019. "Pattern Terrazzo Floors." Accessed February 8, 2024. https://www.aectual.com/architectural-products/flooring/variants/pattern-terrazzo/intro.

Ahlers, Daniel, Florens Wasserfall, Norman Hendrich, and Jianwei Zhang. 2019. "3D Printing of Non-planar Layers for Smooth Surface Generation." In *2019 IEEE 15th International Conference on Automation Science and Engineering (CASE)*, 1737–43. IEEE Press. https://doi.org/10.1109/COASE.2019.8843116.

Allen, Robert J A, and Richard S Trask. 2015. "An Experimental Demonstration of Effective Curved Layer Fused Filament Fabrication Utilising a Parallel Deposition Robot." *Additive Manufacturing* 8: 78–87. https://doi.org/10.1016/j.addma.2015.09.001.

Burger, Joris, Ena Lloret-Fritschi, Fabio Scotto, Thibault Demoulin, Lukas Gebhard, Jaime Mata-Falcón, Fabio Gramazio, Matthias Kohler, and Robert J Flatt. 2020. "Eggshell: Ultra-Thin Three-Dimensional Printed Formwork for Concrete

Structures." *3D Printing and Additive Manufacturing* 7 (2): 48–59. https://doi.org/10.1089/3dp.2019.0197.

Chakraborty, Debapriya, B Aneesh Reddy, and A Roy Choudhury. 2008. "Extruder Path Generation for Curved Layer Fused Deposition Modeling." *Comput. Aided Des.* 40 (2). USA: Butterworth-Heinemann: 235–43. https://doi.org/10.1016/j.cad.2007.10.014.

Chen, Lufeng, Man-Fai Chung, Yaobin Tian, Ajay Joneja, and Kai Tang. 2019. "Variable-Depth Curved Layer Fused Deposition Modeling of Thin-Shells." *Robotics and Computer-Integrated Manufacturing* 57: 422–34. https://doi.org/10.1016/j.rcim.2018.12.016.

Dai, Chengkai, Charlie C L Wang, Chenming Wu, Sylvain Lefebvre, Guoxin Fang, and Yong-Jin Liu. 2018. "Support-Free Volume Printing by Multi-Axis Motion." *ACM Trans. Graph.* 37 (4). New York, NY, USA: Association for Computing Machinery. https://doi.org/10.1145/3197517.3201342.

Etienne, Jimmy, Nicolas Ray, Daniele Panozzo, Samuel Hornus, Charlie C L Wang, Jonàs Martínez, Sara McMains, Marc Alexa, Brian Wyvill, and Sylvain Lefebvre. 2019. "CurviSlicer: Slightly Curved Slicing for 3-Axis Printers." *ACM Trans. Graph.* 38 (4). New York, NY, USA: Association for Computing Machinery. https://doi.org/10.1145/3306346.3323022.

Fang, Guoxin, Tianyu Zhang, Sikai Zhong, Xiangjia Chen, Zichun Zhong, and Charlie C L Wang. 2020a. "Reinforced FDM: Multi-Axis Filament Alignment with Controlled Anisotropic Strength." *ACM Trans. Graph.* 39 (6). New York, NY, USA: Association for Computing Machinery. https://doi.org/10.1145/3414685.3417834.

Gao, Wei, Yunbo Zhang, Devarajan Ramanujan, Karthik Ramani, Yong Chen, Christopher Williams, Charlie Wang, Yung Shin, Song Zhang, and Pablo Zavattieri. 2015. "The Status, Challenges, and Future of Additive Manufacturing in Engineering." *Computer-Aided Design* 69 (August). https://doi.org/10.1016/j.cad.2015.04.001.

Gibson, Ian, David W Rosen, and Brent Stucker. 2009. *Additive Manufacturing Technologies: Rapid Prototyping to Direct Digital Manufacturing*. 1st ed. Springer Publishing Company, Incorporated.

Huang, Bin, and Sarat Singamneni. 2015. "A Mixed-Layer Approach Combining Both Flat and Curved Layer Slicing for Fused Deposition Modelling." *Proceedings of the Institution of Mechanical Engineers, Part B: Journal of Engineering Manufacture* 229 (12): 2238–49. https://doi.org/10.1177/0954405414551076.

Huang, Yijiang, Juyong Zhang, Xin Hu, Guoxian Song, Zhongyuan Liu, Lei Yu, and Ligang Liu. 2016. "FrameFab: Robotic Fabrication of Frame Shapes." *ACM Trans. Graph.* 35 (6). New York, NY, USA: Association for Computing Machinery. https://doi.org/10.1145/2980179.2982401.

Jipa, Andrei, Mathias Bernhard, Benjamin Dillenburger, Nicolas Ruffray, Timothy Wangler, and Robert Flatt. 2017. "SkelETHon Formwork 3D Printed Plastic Formwork for Load-Bearing Concrete Structures." *Blucher Design Proceedings* 3 (12): 345–52. http://dx.doi.org/10.5151/sigradi2017-054.

Keating, Steven, and Neri Oxman. 2013. "Compound Fabrication: A Multi-Functional Robotic Platform for Digital Design and Fabrication." *Robotics and Computer-Integrated Manufacturing* 29 (6): 439–48. https://doi.org/10.1016/j.rcim.2013.05.001.

Kwon, Hyunchul, Martin Eichenhofer, Thodoris Kyttas, and Benjamin Dillenburger. 2019. "Digital Composites: Robotic 3D Printing of Continuous Carbon Fiber-Reinforced Plastics for Functionally-Graded Building Components." In *Robotic Fabrication in Architecture, Art and Design 2018*, edited by Jan Willmann, Philippe Block, Marco Hutter, Kendra Byrne, and Tim Schork, 363–76. Cham: Springer International Publishing.

Leschok, Matthias, and Benjamin Dillenburger. 2019. "Dissolvable 3DP Formwork. Water-Dissolvable 3D Printed Thin-Shell Formwork for Complex Concrete Components." In *ACADIA 19: Ubiquity and Autonomy. Proceedings of the 39th Annual Conference of the Association for Computer Aided Design in Architecture*, edited by Kory Bieg, Danelle Briscoe, and Clay Odom, 188–97. Association for Computer Aided Design in Architecture (ACADIA).

Lim, Sungwoo, Richard A Buswell, Philip J Valentine, Daniel Piker, Simon A Austin, and Xavier De Kestelier. 2016. "Modelling Curved-Layered Printing Paths for Fabricating Large-Scale Construction Components." *Additive Manufacturing* 12: 216–30. https://doi.org/10.1016/j.addma.2016.06.004.

Llewellyn-Jones, Thomas, Robert Allen, and Richard Trask. 2016. "Curved Layer Fused Filament Fabrication Using Automated Toolpath Generation." *3D Printing and Additive Manufacturing* 3 (4): 236–43. https://doi.org/10.1089/3dp.2016.0033.

Ma, Z, A Walzer, C Schumacher, R Rust, F Gramazio, M Kohler, and M Bächer. 2020. "Designing Robotically-Constructed Metal Frame Structures." *Computer Graphics Forum* 39 (2): 411–22. https://doi.org/10.1111/cgf.13940.

Mitropoulou, Ioanna, Mathias Bernhard, and Benjamin Dillenburger. 2020. "Print Paths Key-Framing: Design for Non-Planar Layered Robotic FDM Printing." In *Symposium on Computational Fabrication*. SCF '20. New York, NY, USA: Association for Computing Machinery. https://doi.org/10.1145/3424630.3425408.

Mitropoulou, Ioanna, Amir Vaxman, Olga Diamanti, and Benjamin Dillenburger. 2024. "Fabrication-Aware Strip-Decomposable Quadrilateral Meshes." *Computer-Aided Design* 168 (March). Elsevier BV: 103666. https://doi.org/10.1016/j.cad.2023.103666.

Pan, Yayue, Chi Zhou, Yong Chen, and Jouni Partanen. 2014. "Multitool and Multi-Axis Computer Numerically Controlled Accumulation for Fabricating Conformal Features on Curved Surfaces." *Journal of Manufacturing Science and Engineering* 136 (3). https://doi.org/10.1115/1.4026898.

Pelzer, Lukas, and Christian Hopmann. 2021. "Additive Manufacturing of Non-Planar Layers with Variable Layer Height." *Additive Manufacturing* 37: 101697. https://doi.org/10.1016/j.addma.2020.101697.

Pottmann, Helmut, Alexander Schiftner, Pengbo Bo, Heinz Schmiedhofer, Wenping Wang, Niccolo Baldassini, and Johannes Wallner. 2008. "Freeform Surfaces from Single Curved Panels." *ACM Trans. Graph.* 27 (3). New York, NY, USA:

Association for Computing Machinery: 1–10.
https://doi.org/10.1145/1360612.1360675.

Sarakinioti, Maria-Valentini, Thaleia Konstantinou, Michela Turrin, Martin Tenpierik, Roel Loonen, Marie L de Klijn-Chevalerias, and Ulrich Knaack. 2018. "Development and Prototyping of an Integrated 3D-Printed Façade for Thermal Regulation in Complex Geometries." *Journal of Facade Design and Engineering* 6 (2): 29–40. https://doi.org/10.7480/jfde.2018.2.2081.

Schüller, Christian, Roi Poranne, and Olga Sorkine-Hornung. 2018. "Shape Representation by Zippables." *ACM Trans. Graph.* 37 (4). New York, NY, USA: Association for Computing Machinery. https://doi.org/10.1145/3197517.3201347.

Takezawa, Masahito, Takuma Imai, Kentaro Shida, and Takashi Maekawa. 2016. "Fabrication of Freeform Objects by Principal Strips." *ACM Trans. Graph.* 35 (6). New York, NY, USA: Association for Computing Machinery. https://doi.org/10.1145/2980179.2982406.

Tam, Kam-Ming Mark, and Caitlin T Mueller. 2017. "Additive Manufacturing Along Principal Stress Lines." *3D Printing and Additive Manufacturing* 4 (2): 63–81. https://doi.org/10.1089/3dp.2017.0001.

Taseva, Yoana, Nik Eftekhar, Hyunchul Kwon, Matthias Leschok, and Benjamin Dillenburger. 2020. "Large-Scale 3D Printing for Functionally-Graded Facade." In *Anthropocene, Proceedings of the 25th International Conference of the Association for Computer-Aided Architectural Design Research in Asia (CAADRIA) 2020*, 1:183–92. Hong Kong.

Ulu, Erva, Emrullah Korkmaz, Kubilay Yay, O Burak Ozdoganlar, and Levent Burak Kara. 2015. "Enhancing the Structural Performance of Additively Manufactured Objects Through Build Orientation Optimization." *Journal of Mechanical Design* 137 (11). https://doi.org/10.1115/1.4030998.

Verhoeven, Floor, Amir Vaxman, Tim Hoffmann, and Olga Sorkine-Hornung. 2022. "Dev2PQ: Planar Quadrilateral Strip Remeshing of Developable Surfaces." *ACM Trans. Graph.* 41 (3). New York, NY, USA: Association for Computing Machinery. https://doi.org/10.1145/3510002.

Wu, Chenming, Chengkai Dai, Guoxin Fang, Yong-Jin Liu, and Charlie C L Wang. 2017. "RoboFDM: A Robotic System for Support-Free Fabrication Using FDM." In *2017 IEEE International Conference on Robotics and Automation (ICRA)*, 1175–80. https://doi.org/10.1109/ICRA.2017.7989140.

Wu, Rundong, Huaishu Peng, François Guimbretière, and Steve Marschner. 2016. "Printing Arbitrary Meshes with a 5DOF Wireframe Printer." *ACM Trans. Graph.* 35 (4). New York, NY, USA: Association for Computing Machinery. https://doi.org/10.1145/2897824.2925966.

Zhang, Tianyu, Guoxin Fang, Yuming Huang, Neelotpal Dutta, Sylvain Lefebvre, Zekai Murat Kilic, and Charlie C L Wang. 2022. "S3-Slicer: A General Slicing Framework for Multi-Axis 3D Printing." *ACM Trans. Graph.* 41 (6). New York, NY, USA: Association for Computing Machinery. https://doi.org/10.1145/3550454.3555516.